%% file: main_arxiv.tex
\documentclass{article}

\usepackage{microtype}
\usepackage{graphicx}
\usepackage{subfigure}
\usepackage{booktabs} 

\usepackage{hyperref}

\usepackage[accepted]{icml2024_arxiv}
\input{macros}

\usepackage{amsmath}
\usepackage{amssymb}
\usepackage{mathtools}
\usepackage{amsthm}

\usepackage[capitalize,noabbrev]{cleveref}

\theoremstyle{plain}
\newtheorem{theorem}{Theorem}[section]

\theoremstyle{definition}
\newtheorem{definition}[theorem]{Definition}

\theoremstyle{remark}

\usepackage[textsize=tiny]{todonotes}

\icmltitlerunning{The Limitations of Model Retraining in the Face of Performativity}

\begin{document}

\twocolumn[
\icmltitle{The Limitations of Model Retraining in the Face of Performativity}



\icmlsetsymbol{equal}{*}

\begin{icmlauthorlist}
\icmlauthor{Anmol Kabra}{equal,yyy}
\icmlauthor{Kumar Kshitij Patel}{equal,yyy}
\end{icmlauthorlist}

\icmlaffiliation{yyy}{Toyota Technological Institute at Chicago, United States}

\icmlcorrespondingauthor{Anmol Kabra}{anmol@ttic.edu}
\icmlcorrespondingauthor{Kumar Kshitij Patel}{kkpatel@ttic.edu}

\icmlkeywords{Distribution shifts, Performative prediction, Algorithmic decision-making}

\vskip 0.3in
]



\printAffiliationsAndNotice{\icmlEqualContribution} 

\begin{abstract}
We study stochastic optimization in the context of performative shifts, where the data distribution changes in response to the deployed model.
We demonstrate that naive retraining can be provably suboptimal even for simple distribution shifts.
The issue worsens when models are retrained given a finite number of samples at each retraining step.
We show that adding regularization to retraining corrects both of these issues, attaining provably optimal models in the face of distribution shifts.
Our work advocates rethinking how machine learning models are retrained in the presence of performative effects.
\end{abstract}

\section{Introduction}
\label{sec:introduction}
\input{ICML_src/introduction}

\section{Retraining Fails for Simple Distribution Shifts}
\label{sec:retraining}
\input{ICML_src/retraining}

\section{Empirical Retraining Fails to Converge even with Infinite Samples}
\label{sec:retraining_finite}
\input{ICML_src/retraining_finite}

\section{Regularization Mitigates the Perils of Retraining }
\label{sec:reguilarization}
\input{ICML_src/regularization}

\section{Related Work}
\label{sec:related_work}
\input{ICML_src/related}

\section{Conclusion}
\label{sec:conclusion}
\input{ICML_src/conclusion}

\section*{Acknowledgements}

We thank Saba Ahmadi and Haifeng Xu for insightful discussions on strategic classification.

\bibliography{refs}
\bibliographystyle{icml2024}

\newpage
\appendix
\onecolumn
\input{ICML_src/appendix}

\end{document}

%% file: macros.tex
\input{global_macros}

\DeclareMathOperator*{\PR}{PR}
\newcommand{\PO}{\textrm{PO}\xspace}
\newcommand{\PS}{\textrm{PS}\xspace}
\newcommand{\Stat}{\textrm{Stat}\xspace}

\newcommand{\ignore}[1]{}

%% file: global_macros.tex
\usepackage[utf8]{inputenc}
\usepackage{xspace}

\usepackage{centernot}
\usepackage{mathtools}
\usepackage{bbm}
\usepackage{bm}

\usepackage{xparse}
\ExplSyntaxOn

\NewDocumentCommand{\definealphabet}{mmmm}
 {
  \int_step_inline:nnn { `#3 } { `#4 }
   {
    \cs_new_protected:cpx { #1 \char_generate:nn { ##1 }{ 11 } }
     {
      \exp_not:N #2 { \char_generate:nn { ##1 } { 11 } }
     }
   }
 }

\ExplSyntaxOff

\definealphabet{bb}{\mathbb}{A}{Z}
\definealphabet{cal}{\mathcal}{A}{Z}
\definealphabet{bf}{\bm}{a}{z}
\definealphabet{bf}{\bm}{A}{Z}
\newcommand{\bfzero}{\bm{0}}

\newcommand{\bfmu}{\bm{\mu}}

\newcommand{\bftheta}{\bm{\theta}}
\newcommand{\bfSigma}{\bm{\Sigma}}

\newcommand{\inparens}[1]{\left( #1 \right)}
\newcommand{\inbraces}[1]{\left\{ #1 \right\}}
\newcommand{\inbraks}[1]{\left[ #1 \right]}
\newcommand{\set}[1]{\inbraces{#1}}
\renewcommand{\empty}{\varnothing}

\newcommand{\eps}{\varepsilon}

\newcommand{\defeq}{\triangleq}
\newcommand{\given}{\; \mid \;}

\newcommand{\norm}[1]{\left\lVert #1 \right\rVert}
\newcommand{\dotprod}[2]{\left\langle #1, #2 \right\rangle}

\DeclareMathOperator*{\argmin}{arg\,min}

\newcommand{\gradeval}[2]{\left. #1 \right|_{#2}}

\newcommand{\Esymb}{\mathbb{E}}

\DeclareMathOperator*{\ExpOp}{\Esymb}

\newcommand{\ex}{\ExpOp}

\newcommand{\simiid}{\overset{\mathrm{iid}}{\sim}}

\newcommand{\normal}{\mathcal{N}}

\newcommand{\round}[2]{{#1}^{(#2)}}



%% file: ICML_src/introduction.tex
A major challenge in deploying machine learning models for decision-making is to prevent deteriorating model performance when humans adjust their behavior in response to deployed models.
Conventionally, machine learning models are optimized to generalize to a fixed data distribution.
With humans (or agents) in the loop who actively influence the data distribution, the assumption of a fixed data distribution is at best naive and at worst catastrophically misleading.

Reactivity of agents to predictions about them has been extensively studied in economics, public policy, and sociology.
In the context of machine learning, such predictions have been called \textit{performative}~\cite{perdomo2020performative} as they induce a change in the data distribution.
\citet{perdomo2020performative} introduced the \textit{Performative Risk} to characterize the objective in the presence of performativity:
\begin{align}
\label{eq:PR}
    \PR(\bftheta) &\coloneqq \ex_{\bfz \sim \calD(\bftheta)} \inbraks{ \ell(\bfz; \bftheta) } \enspace.
\end{align}

Here, the loss function $\ell$ is defined for model parameters $\bftheta$ and data samples $\bfz \in \calZ$, such that the samples $\bfz$ are drawn from data distribution $\calD(\bftheta)$ obtained from specified map $\calD : \Theta \to \Delta(\calZ)$.
The hardness of optimizing the objective in \eqref{eq:PR} comes from the fact that even when $\ell(\bfz; \cdot)$ is a convex function, the Performative Risk function, $\PR(\cdot)$, might be non-convex in parameters $\bftheta$.
When there are no performative effects of the model, i.e., when the model does not influence the data distribution, we recover the usual population risk, 
\begin{align}\label{eq:L}
    L(\bftheta) &= \ex_{\bfz \sim \calD} \inbraks{ \ell(\bfz; \bftheta) }\enspace.
\end{align}
In practice, a common strategy for combating such distribution shifts caused by performativity is to \textbf{retrain} the machine learning model at regular intervals on recently collected data samples.
The retraining procedure for time steps $t \in [T]$, referred to as \textit{Repeated Risk Minimization} (R-RM), can be described as follows:
\begin{align}\label{eq:RRM}
    \bftheta_{t} \in \argmin_{\bftheta} \ex_{z \sim \calD(\bftheta_{t-1}) } \inbraks{ \ell(\bfz; \bftheta) } \enspace.
\end{align}
Retraining might not converge without assumptions on the distribution shift function $\calD(\cdot)$, but when it does, we will refer to its \textit{fixed point} as follows, 
 \begin{align}\label{eqn:pr_ps_def}
    \bftheta_\PS &\in \Theta_\PS \coloneqq \argmin_{\bftheta} \ex_{\bfz \sim \calD(\bftheta_\PS)} [\ell(\bfz; \bftheta)]\enspace.
\end{align}

In existing literature, the $\Theta_\PS$ solutions are called the \textit{Performatively Stable} solutions \citep{perdomo2020performative, miller2021outside}.\footnote{We argue in this paper that the ``Performatively Stable'' nomenclature can be misleading as other algorithms can be used to optimize \eqref{eq:PR}.
Consequently, we refer to $\Theta_\PS$ as the fixed points of retraining (\Cref{eq:RRM}).}
Several works, starting from \citet{perdomo2020performative}, have tried to characterize (i) the assumptions on the distribution shift $\calD(\cdot)$ and the loss function $\ell$, which guarantee that retraining/R-RM converges as well as (ii) the rate of this convergence.
While these works provide some theoretical support for using re-training in practice, in this paper, we claim that under very simple distribution shifts, retraining fails catastrophically, in that even if it converges, the quality of the solution, i.e., $\bftheta_\PS$ is much poorer compared to the optimal solution under performative shifts, i.e., 
\begin{align}
\label{eqn:pr_po_def}
    \bftheta_\PO &\in \Theta_\PO \coloneqq \argmin_{\bftheta} \PR(\bftheta)\enspace .
\end{align}
The solutions $\Theta_\PO$ are called \textit{Performatively Optimal}, as they are minimizers for the possibly non-convex function $\PR(\cdot)$.
Because of this non-convexity, we can not hope to recover performatively optimal solutions in every setting~\cite{perdomo2020performative,miller2021outside,mendler2020stochastic,jagadeesan2022regret}. In fact, as we show in the next section, even when $\PR(\cdot)$ is convex, it turns out that for seemingly simple distribution shifts, $\bftheta_\PS$ and $\bftheta_\PO$ may vastly differ~\cite{miller2021outside}.   

%% file: ICML_src/retraining.tex
Before we state our negative result, we will define the following additional solution concept extending the notion of stationarity to our problem,
\begin{align*}
    \Theta_\Stat &\coloneqq \set{ \bftheta \given  \nabla \PR (\bftheta)  = \bfzero }
\end{align*}

Generally, $\Theta_\PO \subseteq \Theta_\Stat$ for any problem, but when $\PR$ is convex, $\Theta_\PO = \Theta_\Stat$. When $\ell$ is strongly-convex in $\bftheta$, $\Theta_\PS$ is a singleton set. Likewise, if $\PR$ is strongly-convex, $\Theta_\PO = \Theta_\Stat$ is a singleton set. In this section, we present two performative prediction problems on the extremes: one where $\Theta_\PO = \Theta_\PS = \Theta_\Stat$, implying naive re-training would succeed, and one where $\Theta_\PS \cap \Theta_\Stat = \empty$, implying retraining would fail to recover a solution in $\Theta_\PO$. We will consider the following class of \textit{linear shifts} for both examples.

\begin{definition}[Linear shifts~\cite{miller2021outside}]
Let $\bfSigma : \Theta \to \bbR^{d \times d}$ and $\bfmu : \Theta \to \bbR^d$ be linear maps, and $\bfz_0 \sim \calD_0$ be such that $\ex[\bfz_0] = \bfzero,\  \ex [\bfz_0 \bfz_0^\top] = \bfI$.
For $\bfSigma_0 \in \bbR^{d \times d}, \bfmu_0 \in \bbR^d$ such that $\bfSigma_0 + \bfSigma(\bftheta)$ is full-rank for all $\bftheta$, $\calD(\bftheta)$ is a linear shift if the random variable $\bfz_{\bftheta} \sim \calD(\bftheta)$ follows the law:
\begin{align*}
    \bfz_{\bftheta} &\coloneqq (\bfSigma_0 + \bfSigma(\bftheta)) \bfz_0 + \bfmu_0 + \bfmu(\bftheta).
\end{align*}    
\end{definition}
We first note the following positive result about retraining and $\bftheta_{\PS}$ for a simple regression problem.
\begin{theorem}[Mean shifts]
\label{thm:prob_where_ps_po_same}
    Let $\calD(\bftheta)$ be a linear shift over $\bfz \in \bbR^d$ with $\bfSigma(\bftheta) = \bf0 \in \bbR^{d \times d}$ and $\norm{\bfmu}_\ast < 1$ (the largest singular value of linear map $\bfmu$).
    Let $\bfA$ be a positive-definite matrix defining the loss function $\ell(\cdot)$ as the Mahalanobis distance $\ell(\bfz; \bftheta) = \frac{1}{2} \norm{\bftheta - \bfz}_{\bfA}^2$.
    Then, $\Theta_\PO = \Theta_\PS = \Theta_\Stat$.
\end{theorem}
The above example highlights that when there are only \textit{mean shifts}, then retraining is effective. But as it turns out considering even slightly more general distribution shifts, which allow performativity in the co-variance of the data, already breaks the above result. In particular, we can prove the following result in the scalar case.

\begin{theorem}[Linear shifts, scalar setting]
\label{thm:prob_where_ps_stat_no_intersection}
    Let $\bfz, \bftheta \in \bbR$ (i.e., scalars), be denoted as $z, \theta$.
    Let $\calD(\theta)$ be a linear shift defined as $z_{\theta} = (\sigma_0 + \sigma \theta) z_0 + \mu_0 + \mu \theta$ for some $\sigma_0, \sigma > 0$ and $\mu_0, \mu \in \bbR$ where $\mu < 1$. Let $\ell(z; \theta)$ be the squared loss $\ell(z; \theta) = \frac{1}{2} (\theta - z)^2$. Then,
    \begin{align*}
        \PR(\theta_{\PS}) - \PR(\theta_{\PO}) = \PR(\theta_{\PO}) \cdot \frac{\sigma^2}{(1-\mu)^2} > 0 \enspace .
    \end{align*}
\end{theorem}
The above theorem generalizes the previous theorem in the scalar setting when $\bfA = \bfI$. We note that if there is a co-variance shift, i.e., $\sigma>0$, then there is a non-zero gap between the performative risks of retraining and the optimal performative risk.

In the next section, we will see that not only is the fixed point of retraining a bad solution under certain shifts but implementing retraining with finite samples can lead to poor convergence or no convergence at all. 

    

%% file: ICML_src/retraining_finite.tex
In practice, it is impossible to implement re-training as discussed in equation \eqref{eq:RRM}. In real settings, at each time step $t$, the re-training algorithm only has access to finite samples from the distribution at that time step, i.e., $S_t\sim \mathcal{D}(\bftheta_{t-1})^{N_t}$. With this in mind, we can state the following empirical retraining algorithm, which we will refer to as \textit{Repeated Empirical Risk Minimization} (R-ERM). For time $t\in[T]$ (initialized at $\bftheta_0$) we have,
\begin{align}\label{eq:RERM}
    \bftheta_{t} &\in \argmin_{\bftheta} \frac{1}{N_{t}} \sum_{i\in[N_t]} \ell(\round{\bfz}{t}_i; \bftheta)\enspace.
\end{align}

It is well known \cite{perdomo2020performative} that repeated risk minimization in \eqref{eq:RRM} converges to $\bftheta_\PS$ under reasonable assumptions about the loss and the distribution shift. For instance, we recall the following result that assumes the distribution shifts are Lipshitz-smooth in $\bftheta$.
\begin{theorem}[\citet{perdomo2020performative}]
    Let loss function $\ell$ be $\gamma$-strongly-convex in $\bftheta$ and $\beta_z$-smooth\footnote{%
    Note that $\ell(\bfz; \bftheta)$ is $\beta_z$-smooth in $\bfz$ if for all $\bftheta \in \Theta, \bfz_1, \bfz_2 \in \calZ$, we have $\norm{ \nabla_{\bftheta} \ell(\bfz_1; \bftheta) - \nabla_{\bftheta} \ell(\bfz_2; \bftheta) } \leq \beta_z \norm{\bfz_1 - \bfz_2}$.%
    }%
    in $\bfz$.
    Let $\calD(\cdot)$ be $\eps$-sensitive, i.e., for all $\bftheta, \bftheta' \in \Theta$,
    $W_1 (\calD(\bftheta), \calD(\bftheta')) \leq \eps \norm{\bftheta - \bftheta'}_2$, where $W_1$ is the Wasserstein-1 distance. If $\eps < \frac{\gamma}{\beta_z}$, then R-RM converges to $\bftheta_\PS$ at a linear rate.
\end{theorem}

A natural question is, then, does R-ERM also converge\footnote{%
We say that R-ERM converges to $\bftheta_\PS$ if $\lim_{t \to \infty} \ex \norm{\bftheta_t - \bftheta_\PS}^2 = 0$---here the expectation is over randomness due to sampling $\round{\bfz}{<t}_i$ in iterations $[t]$.%
} to $\bftheta_\PS$ in the same regime?
We find that, even for simple mean shifts, R-ERM does not converge to $\bftheta_\PS$ if $N_t$ is constant throughout iterations $t$. 
\begin{theorem}[Mean shifts, finite samples]
\label{thm:Repeated-0-ERM__diverges_from_PS__constant_samples}
    Let $\mathcal{D}(\bftheta)$ be a linear shift over $\bfz \in \bbR^{d}$ with $\bfmu(\bftheta) = \mu\bftheta$, $\Sigma_0 = \sigma_0 \bfI_d$ and $\Sigma(\bftheta) = \bf0$, and parameters $\bftheta_0$ be initialized to $\bfzero$.
    Then the parameters $\bftheta_T$ obtained using R-ERM~\eqref{eq:RERM} with constant samples $N_t = N$ in all iterations $t \in [T]$ satisfy:
    \begin{align*}
        \ex \norm{\bftheta_T - \bftheta_\PS}_2^2 &= \norm{\bfmu_0}^2 \frac{\mu^{2T}}{(1 - \mu)^2} + \frac{d \sigma_0^2}{N} \cdot \frac{1 - \mu^{2T}}{1 - \mu^2}\enspace.
    \end{align*}
    In particular, R-ERM does not converge to $\bftheta_\PS$ even when $T\to\infty$.
\end{theorem}
The above theorem is disappointing for naive retraining, showing that the method does not converge to $\bftheta_{\PS}$ even with infinite samples (finite $N$, but infinite $T$). Along with the observations in the last section, this prompts us to design a regularized retraining procedure to rectify the shortcomings of naive retraining.

%% file: ICML_src/regularization.tex
An informed reader may have already guessed that both the issues of fixed-point discrepancy and finite sample errors we have mentioned so far might be corrected with appropriate regularization. In this section, we will show how to regularize retraining appropriately. We define the following regularized retraining, which we refer to as \textit{Regularized Repeated Risk Minimization} (Reg-R-RM). For all $t\in[T]$ and initialization $\bftheta_0$,
\begin{align}\label{eq:RRRM}
    \hspace{-0.8em}\bftheta_{t} &\in \argmin_{\bftheta} \ex_{\bfz \sim \calD(\bftheta_{t-1})} [\ell(\bfz; \bftheta)] + \lambda_{t-1} R(\bftheta, \bftheta_{t-1})\enspace,
\end{align}
where $R(\bftheta, \bftheta_{t-1})$ is a strongly-convex function of $\bftheta$ and $\{\lambda_{t-1}\}_{t\in[T]}$ is the sequence of regularization parameters. In particular, the above algorithm slows down retraining, preventing drastic changes between time steps even when distribution shifts happen. We first show that in the co-variance shift example discussed in Theorem \ref{thm:prob_where_ps_stat_no_intersection}, simple $L_2$-regularization corrects the fixed point discrepancy.

\begin{theorem}[Covariance shift, with regularization]
\label{thm:prob_where_ps_stat_no_intersection_regularized}
    Let $\bfz, \bftheta \in \bbR$ (i.e., scalars) be denoted as $z, \theta$.
    Let $\calD(\theta)$ be a linear shift defined as $z_{\theta} = (\sigma_0 + \sigma \theta) z_0 + \mu_0 + \mu \theta$ for some $\sigma_0, \sigma > 0$ and $\mu_0, \mu \in \bbR$ where $\mu < 1$. Let $\ell(z; \theta)$ be the squared loss $\ell(z; \theta) = \frac{1}{2} (\theta - z)^2$. Then Reg-R-RM with, $$R(\theta, \theta_t) = \frac{\theta^2}{2}, \text{ and } \lambda_t = \lambda^\ast = \frac{\sigma (\mu_0 \sigma + (1-\mu) \sigma_0)}{\mu_0 (1-\mu) - \sigma_0 \sigma},$$ for all $t\in[T]$ converges to $\theta_{\PO}$.
\end{theorem}
While the above regularization can not be implemented in practice without a good approximation of problem-dependent parameters, it highlights important qualitative aspects of the regularization one should use in the face of co-variance shifts. Specifically when $\sigma$ is large (w.r.t., $\sigma_0$) then retraining must be \textit{conservative and regularize more}. In practice, $\sigma$ can be estimated by evaluating the empirical variance of the data. We believe it is possible to design an adaptive method to relax the dependence on at least some problem-dependent parameters. We leave this for future work.

Finally, regularization can also correct the finite sample error incurred by R-ERM, as highlighted in the previous section. We first define the following finite sample version of Reg-R-RM: \textit{Regularized Repeated Empirical Risk Minimization} (Reg-R-ERM). For $t\in[T]$, we sample $S_t\sim \mathcal{D}(\bftheta_{t-1})^{N_t}$ and use initialization $\bftheta_0$,
\begin{align}\label{eq:RRERM}
    \hspace{-0.8em}\bftheta_{t} &\in \argmin_{\bftheta} \frac{1}{N_{t}} \sum_{i\in[N_t]} \ell(\round{\bfz}{t}_i; \bftheta)+ \lambda_{t-1} R(\bftheta, \bftheta_{t-1})\enspace,
\end{align}
where $R(\bftheta, \bftheta_{t-1})$ is a strongly-convex function of $\bftheta$ and $\{\lambda_{t-1}\}_{t\in[T]}$ is the sequence of regularization parameters. We can prove the following result for this algorithm. 

\begin{theorem}[Mean shifts, with regularization]
\label{thm:Repeated-0-ERM__diverges_from_PS__constant_samples_regularized}
    Let $\mathcal{D}(\bftheta)$ be a linear shift over $\bfz \in \bbR^{d}$ with $\bfmu(\bftheta) = \mu\bftheta$ (for $\mu<1$), $\Sigma_0 = \sigma_0\bfI_d$ and $\Sigma(\bftheta)=\bfzero$, then we can obtain the following for square loss $\ell(\bfz;\bftheta) = \frac{1}{2}\norm{\bfz-\bftheta}_2^2$,
    \begin{enumerate}
        \item With sample complexity $N_t = \omega(1)$, e.g. $N_t = \log t$, Reg-R-ERM for some fixed $\lambda \ge 0$ and $R(\bftheta, \bftheta_t) = \frac{1}{2} \norm{\bftheta - \bftheta_t}_2^2$ converges to $\bftheta_\PS$.
        
        \item With sample complexity $N_t = \omega(1/t^2)$, e.g. $N_t = 1/t$ or constant $N_t = N$, Reg-R-ERM for $\lambda_t = t$ and $R(\bftheta, \bftheta_t) = \frac{1}{2} \norm{\bftheta - \bftheta_t}_2^2$ converges to $\bftheta_\PS$.

    \end{enumerate}
\end{theorem}
The first part of the theorem implies, that we can correct the finite sample error without regularization if we grow our number of samples over time. However, if we want to keep the sample load balanced across time---something desirable for real applications where data is only collected in fixed intervals---then increasing regularization corrects the finite sample error incurred by R-ERM. 

%% file: ICML_src/related.tex
\paragraph{Performative Prediction}
The framework of performative prediction addresses data distribution shifts caused due to model deployments~\cite{perdomo2020performative,miller2021outside,mendler2020stochastic,hardt2023performative}.
\citet{perdomo2020performative} characterized Performatively Stable and Optimal solutions in various scenarios of performativity and demonstrated when the retraining procedure (R-RM) attains the Performatively Stable solution as the procedure's fixed point.
\citet{miller2021outside} illustrated differences between the two kinds of solutions, and \citet{mendler2020stochastic,ray2022decision,drusvyatskiy2023stochastic} introduced the \textit{Repeated Gradient Descent} procedure to attain the Optimal solution as its fixed point.
Repeated Gradient Descent is effectively a first-order approximation of our proposed Reg-R-ERM (\Cref{eq:RRERM}), which is a more general procedure that does not assume gradient oracle access.

This line of work in performative prediction has broadly focused on attaining the Performatively Stable solution by modifying R-RM~\cite{mendler2020stochastic,drusvyatskiy2023stochastic,brown2022performative}.
Since such solutions can be drastically suboptimal, as shown in Section 2 and \citet{miller2021outside}, our work adapts R-RM with regularization so that the fixed point matches a Performatively Optimal solution.

\paragraph{Strategic Classification}
Recent work in strategic classification investigates expected risk minimization when humans or agents alter behavior to gain an advantage, colloquially called ``gaming''~\cite{hardt2016strategic,tsirtsis2024optimal,ahmadi2022classification}.
This line of work has conventionally studied interactions with best-response agents, who act rationally to gain an advantage against the deployed model.
The framework of performative prediction conveniently models strategic classification against best-response agents in specific settings of agent utility and cost functions~\cite{perdomo2020performative}---our work investigates retraining dynamics in these settings.
There is also much work in studying the negative social pitfalls of repeated risk minimization, especially in the fairness community~\cite{milli2019social,hashimoto2018fairness,liu2018delayed,taori2023data}.

%% file: ICML_src/conclusion.tex
In this paper, we contribute to the ongoing discourse on performativity in learning by highlighting that retraining (despite being immensely popular) has shortcomings in the face of simple distribution shifts. We show that regularization can rectify shortcomings arising from a fixed-point discrepancy or finite sample errors. There are still several open questions:
\begin{itemize}
    \item It is important to characterize for different classes of distribution shifts and losses how the sets $\Theta_{\PO}$, $\Theta_{\PS}$, and $\Theta_{\Stat}$ relate to each other? We address only the tip of the iceberg by highlighting huge gaps between these solution classes for simple regression problems.
    \item It is unclear if, for every problem where retraining converges to some fixed point, a regularization function can rectify its convergence, forcing it to recover solutions from $\Theta_\PS$. Proving this statement one way or the other can guide the design of future retraining approaches.    
    \item As mentioned in Section \ref{sec:reguilarization}, developing non-parametric regularization methods that do not require the knowledge of problem-dependent parameters and can estimate them on the fly would be handy for improving retraining procedures in practice.
    \item Deriving results similar to ours for classification problems and losses would be handy to draw connections with existing literature on strategic classification, which is of utmost importance to modern machine learning applications in the presence of rational agents. 
    \item Finally, there is some literature on understanding performativity in the presence of multiple agents \cite{piliouras2023multi, jin2023performative}, but this area remains largely untackled. In the age of large language models, which continuously improve from users' feedback, it is important to understand how performativity exacerbates issues around data heterogeneity. The distribution shift model proposed by \citet{patel2024limits} for collaborative linear regression can be a good starting place to study such problems.    
\end{itemize}

%% file: ICML_src/appendix.tex
\section{Proof of Theorem \ref{thm:prob_where_ps_po_same}}
\begin{proof}
    We are given that $\bfz_{\bftheta} \sim \calD(\bftheta)$ obeys the following law for some $\bfz_0$ with $\ex [\bfz_0] = \bfzero$ and $\ex [\bfz_0 \bfz_0^\top] = \bfI$:
    \begin{align*}
        \bfz_{\bftheta} &\defeq \bfSigma_0 \bfz_0 + \bfmu_0 + \bfmu(\bftheta).
    \end{align*}

    The loss function is $\ell(\bfz; \bftheta) = \frac{1}{2} \norm{\bftheta - \bfz}_{\bfA}^2$, the Mahalanobis distance.
    Since, $\ell$ is strongly-convex ($\bfA$ is positive-definite), $\Theta_\PS$ is a singleton set.
    
    We can find $\bftheta_\PS$ by setting gradient to zero.
    \begin{align*}
        \bfzero &= \gradeval{ \nabla_{\bftheta} \ex_{\bfz \in \calD(\bftheta_\PS)} [\ell(\bfz; \bftheta)] }{\bftheta = \bftheta_\PS}\\
        &= \ex_{\bfz \in \calD(\bftheta_\PS)} [A (\bftheta_\PS - \bfz)]\\
        &= \ex_{\bfz_0} [ A (\bftheta_\PS - \bfSigma_0 \bfz_0 - \bfmu_0 - \bfmu(\bftheta_\PS) ) ]\\
        &= \bfA \inparens{ (\bfI - \bfmu) \bftheta_\PS - \ex_{\bfz_0} [\bfSigma_0 \bfz_0] - \bfmu_0 }.
    \end{align*}

    Note that $\ker \bfA = \set{\bfzero}$ and $\bfI - \bfmu$ is invertible as $\norm{\bfmu}_\ast < 1$.
    Then, we get that $\bftheta_\PS = (\bfI - \bfmu)^{-1} \bfmu_0$.

    We can find $\bftheta_\Stat$ similarly:
    \begin{align*}
        \bfzero &= \gradeval{ \nabla_{\bftheta} \PR(\bftheta) }{\bftheta = \bftheta_\Stat}\\
        &= \gradeval{ \nabla_{\bftheta} \ex_{\bfz_0} [ \norm{ \bftheta - \bfSigma_0 \bfz_0 - \bfmu_0 - \bfmu(\bftheta) }_A^2 ] }{\bftheta = \bftheta_\Stat}\\
        &= A \cdot \ex_{\bfz_0} \inbraks{ (\bfI - \bfmu)^\top \inparens{ (\bfI - \bfmu) \bftheta_\Stat - \bfSigma_0 \bfz_0 - \bfmu_0 } }\\
        &= A (\bfI - \bfmu)^\top ( (\bfI - \bfmu) \bftheta_\Stat - \bfmu_0 )
    \end{align*}

    Again, since $\ker \bfA = \set{\bfzero}$ and $\bfI - \bfmu$ is invertible, we get that $\bftheta_\Stat = (\bfI - \bfmu)^{-1} \bfmu_0$.

    Hence, $\Theta_\PO = \Theta_\PS = \Theta_\Stat$.
\end{proof}

\section{Proof of Theorem \ref{thm:prob_where_ps_stat_no_intersection}}
\begin{proof}[Proof of \Cref{thm:prob_where_ps_stat_no_intersection}]
    We are given that $z_\theta \sim \calD(\theta)$ obeys the following law for some $z_0$ with $\ex[z_0] = 0$ and $\ex[z_0^2] = 1$:
    \begin{align*}
        z_\theta &\defeq (\sigma_0 + \sigma \theta) z_0 + \mu_0 + \mu \theta.
    \end{align*}
    We can write the following expression for $\PR(\theta)$ by noting the above:
    \begin{align*}
        \PR(\theta) &= \mathbb{E}_{z\sim\mathcal{D}(\theta)}[(z-\theta)^2/2]\enspace,\\
        &= \mathbb{E}_{z_0}[((\sigma_0 + \sigma \theta) z_0 + \mu_0 + \mu \theta - \theta)^2/2]\enspace,\\
        &= \frac{(\sigma_0 + \sigma \theta)^2 + (\mu_0 + \mu \theta - \theta)^2}{2}\enspace,\\
        &= \frac{(\sigma_0 + \sigma \theta)^2 + ((1-\mu)\theta-\mu_0)^2}{2}\enspace,\\
        &= \frac{\sigma_0^2 + \mu_0^2 + (\sigma^2 + (1-\mu)^2) \theta^2 + 2\sigma\sigma_0\theta-2(1-\mu)\mu_0\theta}{2}\enspace,\\
        &= \frac{\sigma_0^2 + \mu_0^2 + (\sigma^2 + (1-\mu)^2) \theta^2 - 2((1-\mu)\mu_0-\sigma\sigma_0)\theta}{2}\enspace,
    \end{align*}
    which we note is a strongly-convex function, implying that $\theta_{\PO}=\theta_{\Stat}$. With the squared loss, $\Theta_\PS$ is a singleton set. We can find $\theta_\PS$ by explicitly using its definition:
    \begin{align*}
        \theta_\PS &= \arg\min_{\theta}\ex_{z\sim \mathcal{D}(\theta_{\PS})}[(z-\theta)^2/2]\enspace,\\
        &= \ex_{z\sim \mathcal{D}(\theta_{\PS})}[z]\enspace,\\
        &= \ex_{z\sim \mathcal{D}(\theta_{\PS})}[(\sigma_0 + \sigma \theta_\PS) z_0 + \mu_0 + \mu \theta_\PS]\enspace,\\
        &= \mu_0 + \mu \theta_\PS\enspace,
    \end{align*}
    which implies that,
    \begin{align*}
        \theta_\PS = \frac{\mu_0}{1-\mu}\enspace.
    \end{align*}
    We can evaluate $\PR(\theta_{\PS})$ as follows,
    \begin{align*}
        \PR(\theta_{\PS}) &= \frac{(\sigma_0 + \sigma\theta_{\PS})^2}{2}\enspace,\\
        &= \frac{\sigma_0^2 + \sigma^2\mu_0^2/(1-\mu)^2 + 2\sigma\sigma_0\mu_0/(1-\mu)}{2}\enspace,\\
        &= \frac{\sigma_0^2}{2} + \frac{\sigma^2\mu_0^2 + 2\sigma\sigma_0\mu_0(1-\mu)}{2(1-\mu)^2}
    \end{align*}
    Similarly, we can find $\theta_\Stat=\theta_{\PO}$:
    \begin{align*}
        0 &= \gradeval{ \nabla_\theta \ex_{z \sim \calD(\theta)} [(\theta - z)^2]  }{\theta = \theta_\Stat}\enspace,\\
        &= \gradeval{ \nabla_\theta \ex_{z_0} [ (\theta - (\sigma_0 + \sigma \theta) z_0 - \mu_0 - \mu \theta)^2 ] }{\theta = \theta_\Stat}\enspace,\\
        &= \ex_{z_0} \inbraks{ (1 - \sigma z_0 - \mu) \cdot ((1 - \sigma z_0 - \mu)\theta_\Stat - \sigma_0 z_0 - \mu_0) }\enspace,\\
        &= (1-\mu) \ex_{z_0} \inbraks{ (1 - \sigma z_0 - \mu) \theta_\Stat - \sigma_0 z_0 - \mu_0 } - \sigma \ex_{z_0} \inbraks{ z_0 \inbraks{ (1 - \sigma z_0 - \mu) \theta_\Stat - \sigma_0 z_0 - \mu_0 } }\enspace,\\
        &= (1-\mu) [ (1-\mu) \theta_\Stat - \mu_0 ] + \sigma^2 \theta_\Stat + \sigma_0 \sigma\enspace,\\
        &= ( (1-\mu)^2 + \sigma^2 ) \theta_\Stat - (1-\mu) \mu_0 + \sigma_0 \sigma\enspace.
    \end{align*}
    Thus we find that, $$\theta_\Stat = \frac{(1-\mu)\mu_0 - \sigma_0\sigma}{\sigma^2 + (1-\mu)^2}\enspace .$$
    Therefore, $\Theta_\PS \cap \Theta_\Stat = \empty$ as long as $\sigma > 0$, and when $\sigma=0$ we note that $\theta_{\PS} = \theta_{\Stat}$. We can evaluate $\PR(\theta_{\PO})$ as follows,
    \begin{align*}
        \PR(\theta_{\PO}) &= \frac{\sigma_0^2 + \mu_0^2}{2} - \frac{((1-\mu)\mu_0 - \sigma\sigma_0)^2}{2(\sigma^2 + (1-\mu)^2)}\enspace,\\
        &= \frac{\sigma_0^2}{2} + \frac{\mu_0^2\sigma^2 - \sigma_0^2\sigma^2 + 2\sigma\sigma_0\mu_0(1-\mu)}{2(\sigma^2 + (1-\mu)^2)}\enspace,\\
        &= \frac{\sigma_0^2(1-\mu)^2 + \mu_0^2\sigma^2 + 2\sigma\sigma_0\mu_0(1-\mu)}{2(\sigma^2 + (1-\mu)^2)}\enspace.
    \end{align*}
    Finally, we can evaluate $\Delta$ for this loss function,
    \begin{align*}
        \Delta &= \PR(\theta_{\PS}) - \PR(\theta_\PO)\enspace,\\
        &= \frac{\mu_0^2\sigma^2+ 2\sigma\sigma_0\mu_0(1-\mu)}{2}\left(\frac{1}{(1-\mu)^2} - \frac{1}{\sigma^2 + (1-\mu)^2}\right) + \frac{\sigma_0^2\sigma^2}{2(\sigma^2 + (1-\mu)^2)}\enspace,\\
        &= \frac{\sigma^2}{2(1-\mu)^2}\cdot\left(\frac{\mu_0^2\sigma^2 + 2\sigma\sigma_0\mu_0(1-\mu) + \sigma_0^2(1-\mu)^2}{\sigma^2 + (1-\mu)^2}\right)\enspace,\\
        &= \frac{\sigma^2}{(1-\mu)^2}\cdot\PR(\theta_{\PO})\enspace .
    \end{align*}
    
\end{proof}


\section{Proof of Theorem \ref{thm:Repeated-0-ERM__diverges_from_PS__constant_samples}}
\begin{proof}
    A simple calculation for $\bftheta_\PS$ gives (c.f., proof of Theorem \ref{thm:prob_where_ps_po_same}):
    \begin{align*}
        \bftheta_\PS &= \frac{\bfmu_0}{1 - \mu}.
    \end{align*}

    We will prove the theorem by unrolling the recursion of RERM. Plugging in $\ell$ and computing the minimizer, the retraining rule is:
        \begin{align*}
            \bftheta_{t} &= \frac{1}{N_t} \sum_{i=1}^{N_t} \round{\bfz}{t}_i \quad \text{where } \round{\bfz}{t}_1, \dots, \round{\bfz}{t}_{N_t} \simiid \bfmu_0 + \mu \bftheta_{t-1} + \normal (\bfzero, \sigma^2_0 \bfI_d)\\
            \implies \bftheta_{t} &= \bfmu_0 + \mu \bftheta_{t-1} + Z_{t} \quad \text{where } Z_{t} \sim \normal \inparens{ \bfzero, \frac{\sigma_0^2}{N_t}\cdot \bfI_d }.
        \end{align*}

    We can now unroll the recursion (assuming w.l.o.g., that $\bftheta_0=0$),
        \begin{align*}
            \bftheta_{t} &= \bfmu_0 + \mu (\bfmu_0 + \mu \bftheta_{t-2} + Z_{t-1}) + Z_{t}\enspace,\\
            &\quad\vdots\\
            &= \bfmu_0 (1 + \mu + \cdots + \mu^{t-1}) + \mu^{t} \bftheta_0 + \inparens{ Z_t + \mu Z_{t-1} + \cdots + \mu^{t-1} Z_1 }\enspace,\\
            &= \bfmu_0 \inparens{ \frac{1 - \mu^{t}}{1 - \mu} } + \sum_{j=1}^{t} \mu^{t-j} Z_j\enspace,\\
            \implies \bftheta_T &= \bfmu_0 \inparens{ \frac{1 - \mu^T}{1 - \mu} } + \sum_{j=1}^{T} \mu^{T-j} Z_j \quad \text{where } Z_j \sim \normal \inparens{ \bfzero, \frac{\sigma^2}{N_j}\cdot \bfI_d }.
        \end{align*}
        We will now calculate the residual error $\norm{ \bftheta_T - \bftheta_\PS }^2$.
        \begin{align*}
            \norm{ \bftheta_T - \bftheta_\PS }^2 &= \norm{ \bfmu \inparens{ \frac{-\mu^T}{1-\mu} } + \sum_{j=1}^{T} \mu^{T-j} Z_j }^2\enspace,\\
            &= \norm{\bfmu_0}^2 \frac{\mu^{2t}}{(1-\mu)^2} + \norm{\sum_{j=1}^{T} \mu^{T-j} Z_j }^2 - 2 \frac{\mu^T}{1-\mu} \cdot \dotprod{\bfmu_0}{ \inparens{ \sum_{j=1}^{T} \mu^{T-j} Z_j } }\enspace,\\
            &= T_1 + T_2 + T_3\enspace.
        \end{align*}

        Note that $\ex [T_1] = \norm{\bfmu_0}^2 \frac{\mu^{2T}}{(1-\eps)^2}$ and $\ex[T_3] = 0$ as $\ex[Z_j] = 0$ for all $j  0$.
        All that remains is to calculate $\ex [T_2]$:
        \begin{align*}
            \ex [T_2] &= \ex \inbraks{ \norm{ Z_{T} + \mu Z_{T-1} + \cdots + \mu^{T} Z_0 }^2 }\enspace,\\
            &= \ex \inbraks{ \norm{Z_{T}}^2 + \mu^2 \norm{Z_{T-1}}^2 + \cdots + \mu^{2T} \norm{Z_0}^2 }\enspace, && (Z_i, Z_j \text{ are independent for } j \neq i)\\
            &= \ex \inbraks{ \sum_{j=1}^{T} \mu^{2(T-j)} \norm{Z_j}^2 }\enspace,\\
            &= \sum_{j=1}^{T} \mu^{2(T-j)} \cdot \frac{d \sigma_0^2}{N_j}\enspace. && (\text{by linearity of expectation})
        \end{align*}

        If $N_j = N$ for all $j \le T$, then $\ex [T_2] = \frac{d \sigma_0^2}{N} (1 + \mu^2 + \cdots + \mu^{2(T-1)}) = \frac{d \sigma_0^2}{N} \cdot \frac{1 - \mu^{2T}}{1 - \mu^2}$.
        Therefore,
        \begin{align*}
            \ex \norm{\theta_T - \theta_\PS}^2 &= \norm{\bfmu_0}^2 \frac{\mu^{2T}}{(1 - \mu)^2} + \frac{d \sigma_0^2}{N} \cdot \frac{1 - \mu^{2T}}{1 - \mu^2}\enspace,\\
            \implies \lim_{T \to \infty} \ex \norm{\theta_T - \theta_\PS}^2 &= \frac{d \sigma_0^2}{N} \cdot \frac{1}{1 - \mu^2}\enspace. && (\text{recall }\mu < 1)
        \end{align*}

        This finishes the proof.

\end{proof}

\section{Proof of Theorem \ref{thm:prob_where_ps_stat_no_intersection_regularized}}
\begin{proof}
    We first recall from the proof of Theorem \ref{thm:prob_where_ps_stat_no_intersection} that, $$\theta_\Stat = \theta_\PO = \frac{ \mu_0(1-\mu) - \sigma_0 \sigma}{ (1-\mu)^2 + \sigma^2}\enspace.$$
With this in mind, we note the following using the fixed point equation for RRRM where we denote using $\hat\theta$ the fixed point,
\begin{align*}
    &\hat\theta =  \arg\min_{\theta}\ex_{z\sim \mathcal{D}(\hat\theta)}[(z-\theta)^2/2] + \frac{\lambda^\ast}{2}\theta^2\enspace,\\
    &\Rightarrow \hat\theta = \frac{\ex_{z\sim \mathcal{D}(\hat\theta)}[z]}{1+\lambda^\ast}\enspace,\\
    &\Rightarrow \hat\theta = \frac{\mu_0 + \mu\hat\theta}{1+\lambda^\ast}\enspace,\\
    &\Rightarrow \hat\theta = \frac{\mu_0}{1-\mu+\lambda^\ast}\enspace,\\
    &\Rightarrow \hat\theta = \frac{\mu_0(\mu_0(1-\mu) - \sigma_0 \sigma)}{\mu_0(1-\mu)^2 - \sigma\sigma_0(1-\mu) + \sigma^2\mu_0 + \sigma\sigma_0(1-\mu)}\enspace,\\
    &\Rightarrow \hat\theta = \frac{ \mu_0(1-\mu) - \sigma_0 \sigma}{ (1-\mu)^2 + \sigma^2} = \theta_{\PO}\enspace.
\end{align*}
This finishes the proof.
\end{proof}

\section{Proof of Theorem \ref{thm:Repeated-0-ERM__diverges_from_PS__constant_samples_regularized}}
\begin{proof}[Proof of \Cref{thm:Repeated-0-ERM__diverges_from_PS__constant_samples}]
    Recall from previous proofs that,
    \begin{align*}
        \bftheta_\PS &= \frac{\bfmu_0}{1 - \mu}.
    \end{align*}

    We prove each of the two theorem statements by unrolling the R-RERM recursion.
    \begin{enumerate}
        \item We will show that R-RERM converges to $\bftheta_\PS$ for $\lambda_t = \lambda \ge 0$ and sample complexity $N_t = \omega(1)$.

        With the regularizer $R(\bftheta, \bftheta_t) = \frac{1}{2} \norm{\bftheta - \bftheta_t}^2$, the R-RERM update is:
        \begin{align*}
            \bftheta_{t} &= \argmin_{\bftheta} \frac{1}{N_{t}} \sum_{i=1}^{N_t} \ell( \round{\bfz}{t}_i; \bftheta) + \frac{\lambda_{t-1}}{2} \norm{\bftheta - \bftheta_{t-1}}^2\enspace, && \text{where } \round{\bfz}{t}_1, \dots, \round{\bfz}{t}_{N_t} \simiid \calD (\bftheta_{t-1})
        \end{align*}

        Plugging in the squared loss, we can calculate the update at iteration $t$:
        \begin{align*}
            \bftheta_{t} &= \frac{ \frac{1}{N_t} \sum_{i=1}^{N_t} \round{\bfz}{t}_i + \lambda_{t-1} \bftheta_{t-1} }{\lambda_{t-1} + 1}\enspace.
        \end{align*}

        Since each sample is drawn from $\bfmu_0 + \mu \bftheta_t + \normal(\bfzero, \sigma^2 \bfI_d)$, the update is simply $$\bftheta_{t} = \frac{\bfmu_0}{\lambda_{t-1} + 1} + \frac{\lambda_{t-1} + \mu}{\lambda_{t-1} + 1} \bftheta_{t-1} + \frac{1}{\lambda_{t-1} + 1} Z_t,$$ where $Z_t \sim \normal \inparens{ \bfzero, \frac{\sigma^2}{N_t} \bfI_d }$.

        We can thus unroll the recursion for $\bftheta_{t} - \bftheta_\PS$:
        \begin{align}
        \label{eqn:repeated_lambda_t_erm__diff_to_ps}
        \begin{split}
            \bftheta_{t} - \bftheta_\PS &= \frac{\lambda_{t-1} + \mu}{\lambda_{t-1} + 1} \inbraks{ \bftheta_{t-1} - \bftheta_\PS + \frac{Z_t}{\lambda_{t-1} + \mu} }\enspace,\\
            &= \frac{\lambda_{t-1} + \mu}{\lambda_{t-1} + 1} \inbraks{ \frac{\lambda_{t-2} + \mu}{\lambda_{t-2} + 1} \inbraks{ \bftheta_{t-2} - \bftheta_\PS + \frac{Z_{t-1}}{\lambda_{t-2} + \mu} } + \frac{Z_t}{\lambda_{t-1} + \mu} }\enspace,\\
            &\vdots\\
            &= \inbraks{ \prod_{i=1}^{t} \frac{\lambda_{i-1} + \mu}{\lambda_{i-1} + 1} } (\bftheta_0 - \bftheta_\PS) + \sum_{i=1}^t \inbraks{ \prod_{j=i}^t \frac{\lambda_{j-1} + \mu}{\lambda_{j-1} + 1} } \cdot \frac{Z_i}{\lambda_{i-1} + \mu}\enspace.
        \end{split}
        \end{align}

        Now, fix hyperparameter $\lambda_t = \lambda \ge 0$.
        The deviation of the $t^{th}$ update from $\bftheta_\PS$ is:
        \begin{align*}
            \bftheta_{t} - \bftheta_\PS &= \inparens{ \frac{\lambda + \mu}{\lambda + 1} }^{t} (\bftheta_0 - \bftheta_\PS) + \sum_{i=1}^t \inparens{ \frac{\lambda + \mu}{\lambda + 1} }^{t+1-i} \cdot \frac{Z_i}{\lambda + \mu}\enspace,\\
            \implies \ex \norm{\bftheta_{t} - \bftheta_\PS}^2 &= \inparens{ \frac{\lambda + \mu}{\lambda + 1} }^{2t}\norm{ \bftheta_0 - \bftheta_\PS }^2 + \sum_{i=1}^t \inparens{ \frac{\lambda + \mu}{\lambda + 1} }^{2(t+1-i)} \cdot \frac{\ex \norm{Z_i}^2 }{(\lambda + \mu)^2}\enspace,\\
            &= \inparens{ \frac{\lambda + \mu}{\lambda + 1} }^{2t} \norm{ \bftheta_0 - \bftheta_\PS }^2 + \sigma^2 d \sum_{i=1}^t \inparens{ \frac{\lambda + \mu}{\lambda + 1} }^{2(t+1-i)} \cdot \frac{1}{N_i (\lambda + \mu)^2}\enspace,\\
            &= T_1 + T_2\enspace.
        \end{align*}
        which follows from the fact that $Z_i \sim \normal \inparens{ \bfzero, \frac{\sigma^2}{N_i} \bfI_d }$ and $Z_i, Z_j$ are independent for $j \neq i$.
        
        As $\eps < 1$, $\lim_{t \to \infty} T_1 = 0$.
        Moreover, if $N_t = \omega(1)$ such as $N_t = \log t$ then $\lim_{t \to \infty} T_2 = 0$.
        Therefore, $\lim_{t \to \infty} \ex \norm{\bftheta_t - \bftheta_\PS}^2 = 0$.
        This shows that R-RERM converges with a growing logarithmic sample complexity schedule.
        
        \item We will show that R-RERM converges to $\bftheta_\PS$ for $\lambda_t = t$ and sample complexity $N_t = \omega(1/t^2)$.
        Recall the recursion for $\bftheta_{t+1} - \bftheta_\PS$ from \Cref{eqn:repeated_lambda_t_erm__diff_to_ps},
        \begin{align*}
            \bftheta_{t} - \bftheta_\PS &= \inbraks{ \prod_{i=1}^{t} \frac{\lambda_{i-1} + \mu}{\lambda_{i-1} + 1} } (\bftheta_0 - \bftheta_\PS) + \sum_{i=1}^t \inbraks{ \prod_{j=i}^t \frac{\lambda_{j-1} + \mu}{\lambda_{j-1} + 1} } \cdot \frac{Z_i}{\lambda_{i-1} + \mu}\enspace.
        \end{align*}

        Now, fix the hyperparameter $\lambda_t = t+1$.
        The deviation of the $t^{th}$ update from $\bftheta_\PS$ is,
        \begin{align*}
            \bftheta_{t} - \bftheta_\PS &= \inbraks{ \prod_{i=1}^t \frac{t + \mu}{t + 1} } (\bftheta_0 - \bftheta_\PS) + \sum_{i=1}^t \inbraks{ \prod_{j=i}^t \frac{j + \mu}{j + 1} } \cdot \frac{Z_i}{i + \mu}\\
            \implies \ex \norm{\bftheta_{t} - \bftheta_\PS}^2 &= \inbraks{ \prod_{i=1}^t \inparens{ \frac{t + \mu}{t + 1} }^2 } \norm{\bftheta_0 - \bftheta_\PS}^2 + \sum_{i=1}^t \inbraks{ \prod_{j=i}^t \inparens{ \frac{j + \mu}{j + 1} }^2 } \cdot \frac{\ex \norm{Z_i}^2}{(i + \mu)^2}\\
            &= \inbraks{ \prod_{i=1}^t \inparens{ \frac{t + \mu}{t + 1} }^2 } \norm{\bftheta_0 - \bftheta_\PS}^2 + \sigma^2 d \sum_{i=1}^t \inbraks{ \prod_{j=i}^t \inparens{ \frac{j + \mu}{j + 1} }^2 } \cdot \frac{1}{N_i (i + \mu)^2}\\
            &= T_1 + T_2.
        \end{align*}
        which follows from the fact that $Z_i \sim \normal \inparens{ \bfzero, \frac{\sigma^2}{N_i} \bfI_d }$ and $Z_i, Z_j$ are independent for $j \neq i$.
        
        As $\eps < 1$, $\lim_{t \to \infty} T_1 = 0$.
        Moreover, if $N_t = \omega(1/t^2)$ such as $N_t = \Theta(1)$ or $N_t = \Theta(1/t)$ then $\lim_{t \to \infty} T_2 = 0$.
        Therefore, $\lim_{t \to \infty} \ex \norm{\bftheta_t - \bftheta_\PS}^2 = 0$.
        This shows that R-RERM converges with a diminishing (or constant) sample complexity schedule.
    \end{enumerate}
    This finishes the proof.
\end{proof}